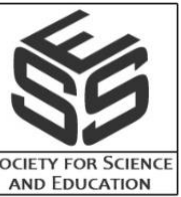

# An Efficient Application of Neuroevolution for Competitive Multiagent Learning

**Unnikrishnan Rajendran Menon**
School of Electrical Engineering, Vellore Institute of Technology,
Vellore, Tamil Nadu, India;

**Anirudh Rajiv Menon**
School of Electronics Engineering, Vellore Institute of Technology,
Vellore, Tamil Nadu, India;

**ABSTRACT**
**Multiagent systems provide an ideal environment for the evaluation and analysis of real-world problems using reinforcement learning algorithms. Most traditional approaches to multiagent learning are affected by long training periods as well as high computational complexity. NEAT (NeuroEvolution of Augmenting Topologies) is a popular evolutionary strategy used to obtain the best performing neural network architecture often used to tackle optimization problems in the field of artificial intelligence. This paper utilizes the NEAT algorithm to achieve competitive multiagent learning on a modified pong game environment in an efficient manner. The competing agents abide by different rules while having similar observation space parameters. The proposed algorithm utilizes this property of the environment to define a singular neuroevolutionary procedure that obtains the optimal policy for all the agents. The compiled results indicate that the proposed implementation achieves ideal behaviour in a very short training period when compared to existing multiagent reinforcement learning models.**

**Keywords:** Genetic Algorithm; NeuroEvolution; Neural Networks; Reinforcement Learning; Multiagent Environment

## INTRODUCTION

Due to the rapid developments in the field of Artificial Intelligence, algorithms must be analyzed and observed constantly. Game environments consisting of many variables with unpredictable behavior serve as ideal platforms for this purpose [1]. The traditional algorithms used to solve these environments utilize multiple search paradigms and mechanisms which take a very long time to find optima [2].

Reinforcement learning (RL) is a field of machine learning, that trains agents to learn the optimal policy to successfully navigate an environment. The agent learns to do so by trying to maximize the cumulative reward it receives, based on the actions it takes during different states in the environment, during its training experiences. Over the past decade, research in the field of Reinforcement Learning (RL) has gained huge popularity due to its extensive applications in control systems, robotics, and other optimization techniques for solving real world problems. For instance, the use of reinforcement learning based AI agents by DeepMind to cool Google

Data Centers led to a 40% reduction in energy spending. The centers are now fully controlled by these agents under the supervision of data center experts [3].

Game environments provide quantified state parameters that accurately represent the agents and other contributing factors for any given time step. These parameters can be normalized into a range compatible with various self-learning algorithms including those that rely on neural networks as their backbone structures. These algorithms return numerically encoded actions back to the environment which are translated into the appropriate changes in states and the associated rewards pertaining to the new state.

Reinforcement algorithms such as Deep Q-Learning have been extensively used to study the behaviour of agents by creating different scenarios using game environments. For instance, Deep Q-networks that process raw pixel data, of the states of a pong environment modified to pit 2 agents against a hard-coded paddle, using convolutional neural networks (CNNs) have been used to explore the development of cooperation between agents with a shared goal [4]. Hence, game environments are a medium to simulate a lot of practically infeasible scenarios for the evaluation and testing of various algorithms as they save time and resources.

These DQN based techniques have been extended to form generic agents that can beat most major Atari games. However, these algorithms could take hours to train properly as it takes some time for the network to process the most relevant features from the raw-pixel data returned by the environment [5].

Genetic algorithms are random heuristic operations that optimize by searching the local neighborhood of the parameters to be improved and promote more effective parameters by imitating the mechanics of natural selection and genetics. In comparison to the traditional algorithms, Genetic Algorithms are fast, robust, and, with modifications, even escape converging to local optima. Their functionality not only includes the solving of general, but also unconventional optimization problems often encountered in artificial intelligence [6,7,8]. Genetic algorithms have been used to model and study cooperative behavior among micro-organisms like bacteria and viruses [9,10]. Such evolution-based computations have played a key role in studying Ant and bee colonies as well [11,12].

NeuroEvolution of Augmenting Topologies (NEAT) algorithm utilizes concepts stemming from genetic evolution to search the space of neural network policies for maximizing the fitness function metric [13]. The fitness value of an agent represents how well it performed in each training episode. While value-based reinforcement learning algorithms deploy single agents that learn progressively, NEAT algorithm uses a population of agents to find the best policies. Offspring topologies are obtained via mutation and crossover procedures on the population's fittest individuals. The population eventually performs well enough to cross the desired fitness threshold. It has been observed that for many Reinforcement Learning applications, the NEAT algorithm outperforms other conventional methods [15]. Neuroevolution based algorithms have thus been functional in modelling intelligent agents that can efficiently adapt to strategy-based games, the design of mobile robots, autonomous vehicles and even control systems in aerospace [16,17,18,19,20].

Most multiagent environments, such as the one used in this paper, involve all the agents interacting with the environment under different rules while having similar observation and action spaces, Consequently, NEAT has proved to be an effective algorithm for multiagent learning [21].

An efficient implementation of Neuroevolution on a predator-prey environment has been done to evaluate this approach in terms of the development of policies such that a collaborative nature arises among the predators and on how reward structure and coordination mechanism affect multiagent evolution [22]. Additionally, neuroevolutionary techniques such as Cooperative Synapse Neuroevolution (CoSyNE) have been developed to solve multiagent versions of the pole balancing problem that involve continuous state spaces [14].

This paper implements a single neuroevolutionary strategy to optimize all the agents involved in an indigenously designed multi-paddle pong environment, propagating only the best performing architectures to promote a competitive learning paradigm. The initialization of a single population for agents with different action spaces contributes to the better performance of the proposed algorithm compared to traditional applications of NEAT, as well as other reinforcement learning based techniques, on multiagent systems.

## PROPOSED MULTIAGENT ENVIRONMENT

The environment comprises of a continuous state space within a window of dimensions $800 \times 800$ pixels. The aim is to obtain paddles of all 4 classes, one for each side of the window, with optimized neural networks backing their actions to prevent any of them from missing the ball, via the proposed algorithm. For any paddle initialized during the training stage, the action and observation spaces will be governed by the side of the window it gets initialized on. The paddles initialized on the top and bottom sides of the window can move left or right, while those on the left and right sides of the window can move in the upward or downward directions.

The ball for every training episode is initialized in the middle of the environment and given a random velocity (whose magnitude and direction are arbitrarily chosen within a continuous range) to study how reward structure and individual genome fitness result in competitive multiagent evolution.

The proposed algorithm involves initializing a single population for every training episode which will include multiple paddles belonging to the 4 classes as shown in Figure 1. As a result, the inputs to the neural network for each paddle must comprise of components from the environment that can be universally calculated and represented for all 4 paddle classes. The dynamic input metrics used are delineated in section 4.1.

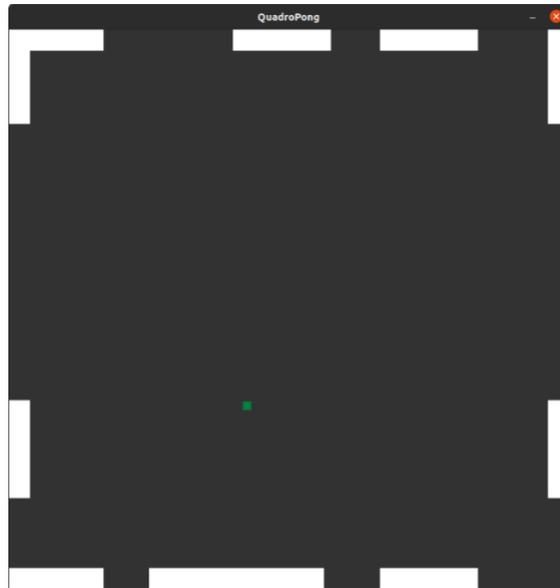

**Figure 1: Population of paddles in proposed environment during training**

## OPTIMIZATION USING NEAT ALGORITHM

Genetic algorithms utilize random heuristic search operations that are mathematically modelled to emulate the technicalities of biological selection and genetics. These algorithms encode samples based on a given set of constraints as individuals that form a population. They promote the best performing individuals by using a randomized yet structured information exchange [23].

NEAT is an evolutionary technique that follows genetic algorithm procedures to optimize neural network architectures [24,25]. This is achieved by initializing a population of artificial neural networks with fully connected input and output neuron layers. Those individuals with the best fitness values are propagated to the subsequent generation. These neural network topologies are then augmented in a probabilistic addition, deletion or modification of hidden neurons, connections, and weights. The evolutionary procedure is carried out till an individual or group of individuals achieves the desired fitness value in a training episode.

The NEAT procedure conducts optimization, on the agents of the proposed environment, by adhering to the following genetic evolution steps:

### Fitness Evaluation

At this stage, the fitness values are calculated for all individuals as actions are conducted by them during the training episode. This is a measure of how well each initialized paddle has performed in the environment. The fitness value in this implementation has been initialized to be 0 for every paddle. Each time a paddle in the population successfully hits the ball, its fitness value receives a large positive increment and for every instant the paddle exists in the environment (does not terminate), it again receives a small positive increment to its fitness function. If the paddle misses the ball, then its fitness value is decremented and is terminated. If the fitness value of any paddle in the population reaches a predefined threshold (which is set to a very high score), then the complete process will be terminated as a desired architecture will have been achieved.

**Selection**
The top 20% of the best performing paddles of the population (according to fitness value) are selected and propagated to the next generation, while the rest are discarded. These selected individuals become part of the next generation as parent neural network architectures.

**Crossover**
This step is analogous to biological reproduction. Here child neural network topologies are obtained from the selected parents of the previous generation. Each offspring has its characteristic architecture derived from a stochastic combination of the genes of its parent agents' architectures. This is done by assigning an innovation number to every gene. The genes of the parent agents with matching innovation numbers are lined up. In case of gene mismatch, the contribution is made by the more fit parent. In case they are equally fit, the gene is inherited from the parents randomly [26]. For the NEAT algorithm, this process is carried out within the network weight vectors to optimize the connection weights that determine the functionality of a network [25].

**Mutation**
Finally, the members of the new population are made to undergo mutation, and this step is responsible for the exploration of the search space as well as escaping local optima. Here the architectures of the newly formed population of paddles are mutated by adding/removing connections, modifications in synaptic weights by small amounts or introducing hidden layer neurons. The mutation probabilities represent the likeliness of a member of the population is to undergo such mutation. For this implementation, the connection add/delete probability is set to 0.5, the node add/delete probability is set to 0.2 and the weight mutate probability and weight replacement probabilities have been set to 0.8 and 0.1 respectively.

Once these procedures have been completed, a new generation of paddles is obtained and made to undergo the fitness evaluation, selection, crossover, and mutation until a member of one of the generations is able to score above the desired threshold fitness value. The promotion of only the fittest few individuals in every generation, ensures the generation of superior offspring leading to competitive training.

## PROPOSED NEAT APPLICATION ON ENVIRONMENT

**Environment and Population Initialization**
- A population of $4n$ paddles is initialized such that each of the 4 paddle classes i.e., the left, right, top, and bottom categories, contain $n$ paddles each.
- A rudimentary neural network is assigned to each individual paddle having 2 neurons in the input layer and 1 neuron in the output layer. The connections and synaptic weights of each neural network are stochastically determined and hence are different from each other for exploration of the search space.
- Each paddle is given the absolute difference between x and y coordinates of the ball and paddle as the input metric to its neural network at every state. This ensures that each agent can perceive changes in the environment as the ball moves around during training.
- The output layer of each neural network returns an action value between $-1$ and $1$ as the activation function used in each layer is the tanh function. The tanh activation function is depicted in Figure 2 and defined below:

$$\tanh(x) = \frac{e^x - e^{-x}}{e^x + e^{-x}} \tag{1}$$

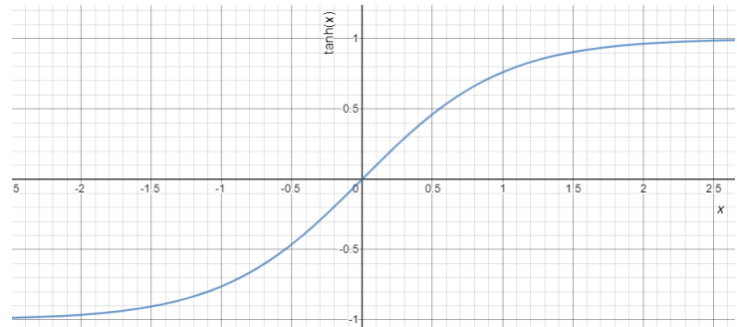

**Figure 2: Tanh Activation Function**

- Based on the category of paddle, this value is interpreted differently. For the top and bottom classes of paddles, a value less than 0 implies a movement to the left and an action value greater than 0 causes it to move right. Similarly, for the right and left classes of paddles, a value less than 0 implies a downward movement and an action value greater than 0 causes it to move upwards. Thus, a common interpretation of the environment as input space is utilized allowing a single population with all types of paddles to train while coexisting.
- At the beginning of each training episode, the ball is initialized at the center of the environment with an arbitrary magnitude and direction of velocity to prevent biased learning.

**Fitness Evaluation**

The fitness metric is a direct indication of how well an individual has performed in the current generation with respect to other individuals of the same generation. The individuals (genomes) are evolved till desirable threshold conditions are met. The fitness values are initialized as 0 for all genomes in a generation which varies as the training episode goes on.

- Every time a paddle successfully hits the ball during the episode, its fitness value is incremented by a magnitude of 10.
- Every time a paddle fails to hit the ball, its fitness value is reduced by a punishment factor having magnitude of 5.

**Training Procedure**
- The environment is initialized with a population of paddles belonging to different classes and a ball with random velocity.
- The paddles are made to play the game and their fitness values are updated at each step as discussed in section 4.2.
- The paddles that fail to hit the ball are not allowed to participate further in the training episode.
- The training episode is terminated if all the paddles on a side of the window fail to hit the ball.

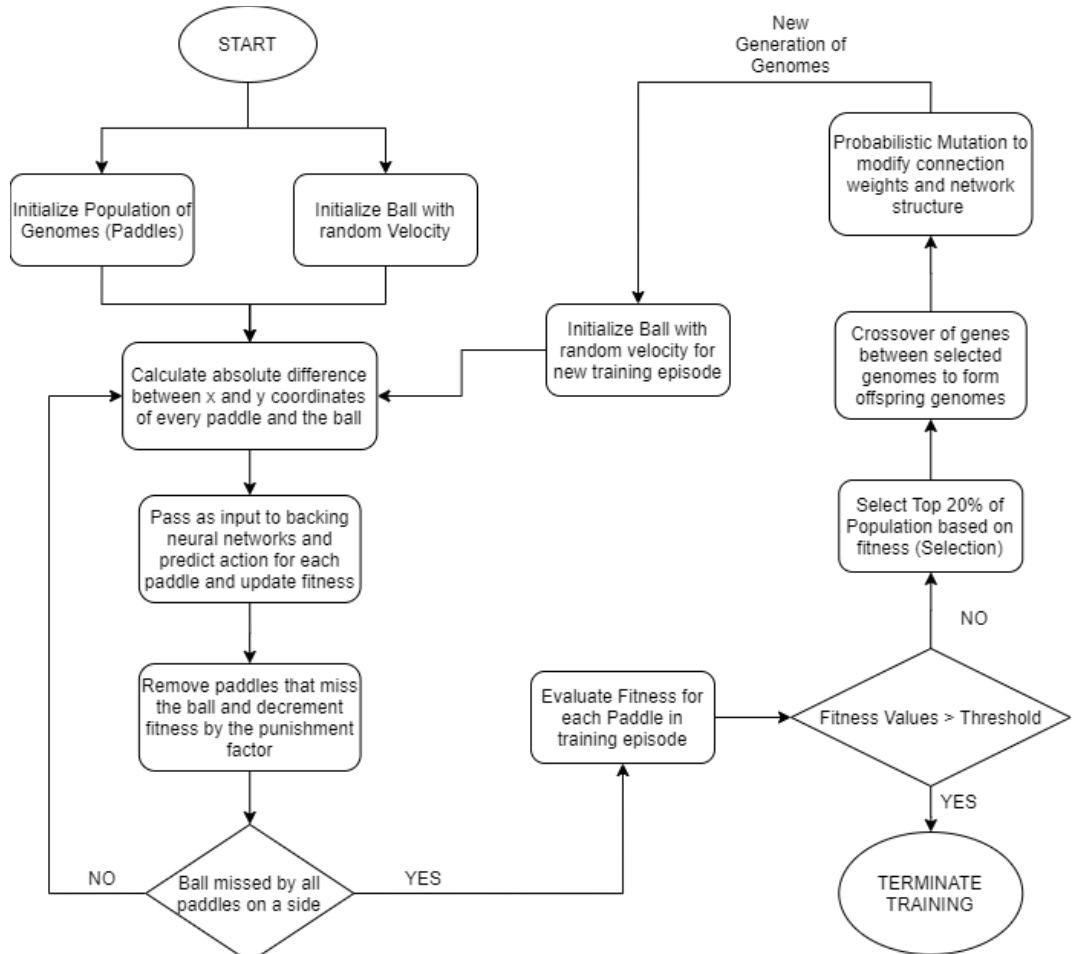

**Figure 3: Flow diagram of proposed system.**

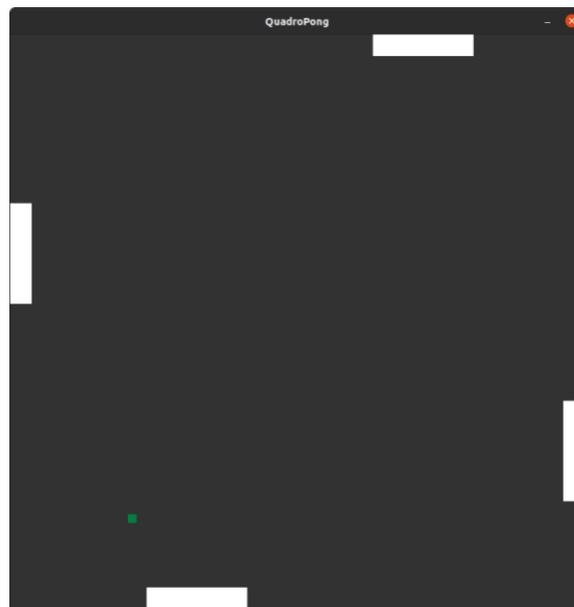

**Figure 4: Trained paddles competing in environment.**

At the end of the training episode, the selection, crossover, and mutation steps are applied to the generation as mentioned in section 3. Thus, a new generation of paddles is obtained. A new training episode is initialized for the new generation and this process is repeated till a population of paddles achieves a fitness value greater than the defined threshold of 100000. Hence a population of agents that successfully plays the game is obtained using only a single evolving population.

## EXPERIMENTAL ANALYSIS

### Population vs Generations

The proposed system was evolved with varying values of the genome population size. The number of generations required by the populations of varying sizes to learn the ideal policy to sustain game is shown in Table 1. Note that sustaining the game requires the paddles to achieve a fitness greater than 25000 as this value was found enough to represent a learned population. The analysis was done with an upper limit of 100 generations for every population size.

The analysis confirms that the general trend is that as the population size is increased, the system achieves a sustained game episode in fewer generations. For a population size of 4, it was observed that the stochastic variations brought about by the NEAT procedure were insufficient to produce optimal neural network architectures in a single population within a reasonable amount of training generations. The population size of 20 was observed to achieve learning within a reasonably short number of generations and hence has been utilized for the subsequent experimental analyses.

However, for population sizes largely exceeding 20, it was observed that it takes much longer to learn the optimal policy. This is attributed to the instances where many paddles on one of the 4 sides of the window hit the ball, resulting in a large number of paddles, pertaining to only one side, achieving a high fitness value. Consequently, the NEAT algorithm propagates offspring networks that tend to perform well on that side. This biased learning paradigm causes these networks to perform poorly when they are spawned on any other sides of the window in future generations. Hence, it takes a very large number of generations for variations in the neural networks, large enough to explore the state space for paddles on other sides, to occur.

Table 1. Variation of number of generations required for training with population size.

| Population | Generations |
|---|---|
| 4 | $\geq 100$ |
| 8 | 46 |
| 16 | 24 |
| 20 | 22 |
| 32 | 85 |
| 40 | $\geq 100$ |

### Analysis of Training Metrics Across Generations

The variation in fitness values with respect to generations as well as the number of agents remaining at the end of each generation were plotted for a population size of 20.

It can be observed from Figure 5 that as the generations progressed, the fitness achieved by the paddles also improved.

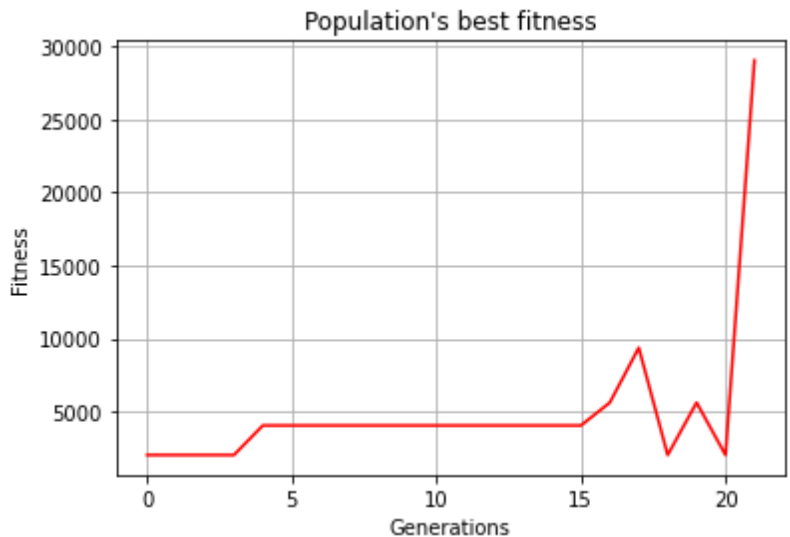

**Figure 5: Best fitness achieved for varying generations.**

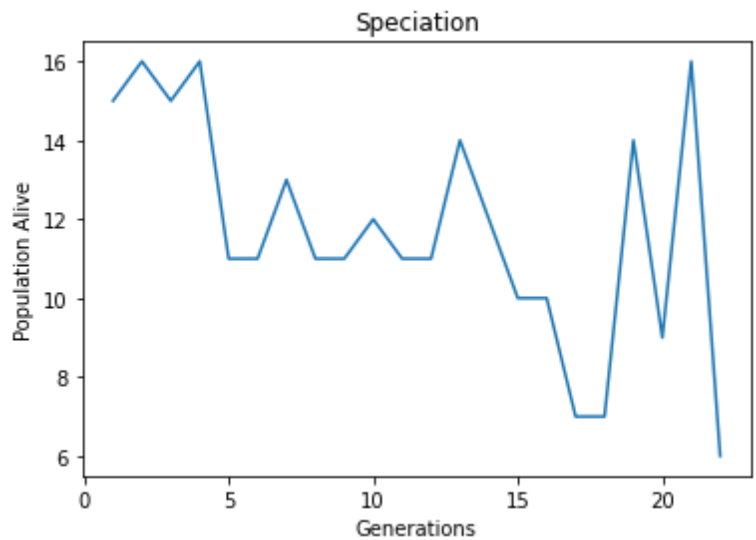

**Figure 6: Number of paddles remaining after every generation.**

Figure 6 draws the inference that as training continues the initial population of 20 paddles converges towards 4 competing paddles that are present on the 4 sides of the window. These remaining paddles represent the neural networks with the best policies to ensure that none of them miss the ball. Since only the top performing neural networks were made to reproduce to generate the subsequent genomes, the resulting trained paddles have achieved their policies because of competition between individuals at every generation.

A deep RL based implementation that trains a single paddle to play against a hard-coded agent, using pixel data as input to a neural network architecture, required 10 million iterations over 6 hours of training on GPU [27]. Another research venture, that utilized Advantage Actor-Critic (A2C), Actor-Critic with Experience Replay (ACER) and Proximal Policy Optimization (PPO) to train a single paddle against a hardcoded agent on the OpenAI Pong environment, observed that these advanced RL techniques required 10 million timesteps across 100 episodes to achieve mean final rewards of 19.7, 20.7 and 20.7 respectively [28].

In comparison, the proposed system trains a configuration of 4 competing paddles in 22 episodes within a few minutes, depending on the initialization of the neural networks in the $1^{st}$ generation of agents. Upon experimentation, it was observed that the system proposed in this paper, for a population of 20 agents per generation take around 20000 timesteps to achieve an infinitely sustaining multiagent game episode, i.e., the agents never lose. Hence, the proposed system trains a multi-agent (paddle agents on each side of the window) competitive pong environment in a much shorter period than what many advanced reinforcement learning algorithms take to train a single paddle agent.

**Analysis of Proposed Algorithm on different Multi-Agent Scenarios**

The proposed implementation of Neuroevolution using only a single population was compared for different multi-agent configurations of pong game i.e., training competing agents on 2 sides of the window and on all 4 sides of the window, the latter case being utilised to detail the proposed system in the sections above. Further, these cases were also compared to the implementation of NEAT on a more traditional pong environment involving only a single paddle. In this scenario, the other three sides of the window were made to be reflective. Figure 7 shows the single and double agent cases implemented on the environment. Table 2 shows the number of generations required by the various multi-agent scenarios to achieve training when initialized with the respective population sizes.

Table 1. Training results for different game scenarios.

| No. of Active Agents | Generations | Population |
|---|---|---|
| 1 | 1 | 4 |
| 2 | 3 | 8 |
| 4 | 22 | 20 |

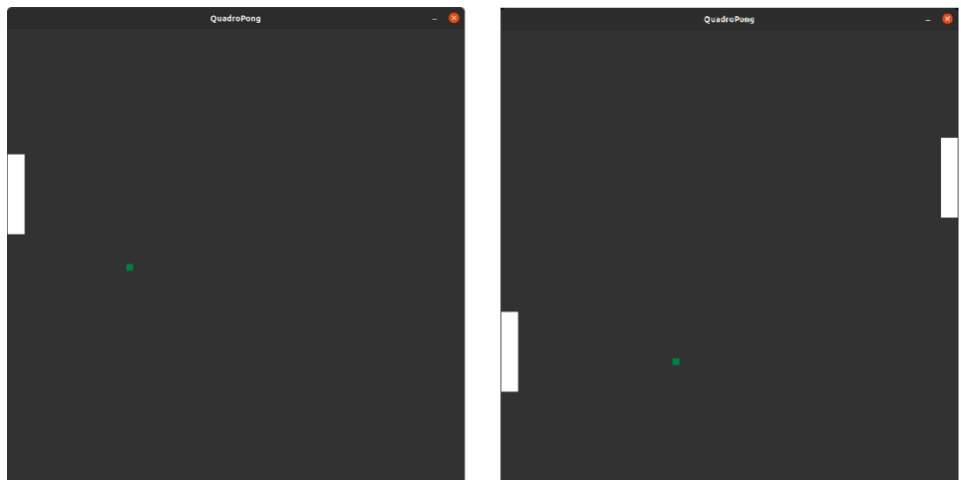

**Figure 7: Application in single and dual agent configurations**

It was observed that for scenarios with fewer competing agents, the proposed system was able to adapt and show intelligent behavior at a much earlier stage compared to cases with more competing agents. This also confirms the idea that the proposed system of initializing a single population for multi-agent learning can be extended to other systems.

## CONCLUSION

Neuroevolutionary algorithms are efficient strategies to bring about competitive and intelligent behaviour in multiagent systems. This paper utilizes NEAT, a neuroevolutionary genetic algorithm which optimizes neural network architectures of agents by promoting competitive breeding and mutation, to train a multi-agent pong environment such that paddle agents on all sides of the game window never fail to hit the ball. The initializing of only a single population for all paddle classes, unlike the traditional NEAT approach to multi-agent problems which utilize separate populations for each agent, demonstrated the ability to adapt in minimal time to develop sufficiently complex learning strategies during the training phase, leading to the conclusion that the research holds many other potential applications. The proposed system was observed to achieve an ideal or infinitely running episode in a much shorter number of time-steps than that required by many advanced Reinforcement Learning algorithms to achieve a sufficiently long, but finite, game episode. The experiments confirmed that the paddle competing mechanism, reward structure and game dynamics are important factors in the evolution of efficient competitive behavior. The work done so far has been compiled into a GitHub repository available here [29].